# Think as Cardiac Sonographers: Marrying SAM with Left Ventricular Indicators Measurements According to Clinical Guidelines

Tuo Liu, Qinghan Yang, Yu Zhang, Rongjun Ge, Yang Chen, Guangquan Zhou

**Abstract.** Left ventricular (LV) indicator measurements following clinical echocardiography guidelines are important for diagnosing cardiovascular disease. Although existing algorithms have explored automated LV quantification, they can struggle to capture generic visual representations due to the normally small training datasets. Therefore, it is necessary to introduce vision foundational models (VFM) with abundant knowledge. However, VFMs represented by the segment anything model (SAM) are usually suitable for segmentation but incapable of identifying key anatomical points, which are critical in LV indicator measurements. In this paper, we propose a novel framework named AutoSAME, combining the powerful visual understanding of SAM with segmentation and landmark localization tasks simultaneously. Consequently, the framework mimics the operation of cardiac sonographers, achieving LV indicator measurements consistent with clinical guidelines. We further present filtered cross-branch attention (FCBA) in AutoSAME, which leverages relatively comprehensive features in the segmentation to enhance the heatmap regression (HR) of key points from the frequency domain perspective, optimizing the visual representation learned by the latter. Moreover, we propose spatial-guided prompt alignment (SGPA) to automatically generate prompt embeddings guided by spatial properties of LV, thereby improving the accuracy of dense predictions by prior spatial knowledge. The extensive experiments on an echocardiography dataset demonstrate the efficiency of each design and the superiority of our AutoSAME in LV segmentation, landmark localization, and indicator measurements. The code will be available at https://github.com/QC-LIU-1997/AutoSAME.

## 1 Introduction

2D echocardiography plays an important role in the measurements of left ventricular (LV) indicators, which facilitate the diagnosis of cardiovascular diseases [1, 2]. At present, most echocardiography societies recommend the biplane Simpson's method to compute LV indicators, such as the end-diastolic length (EDL) and end-systolic length (ESL) of the LV long axis, end-diastolic volume (EDV), end-systolic volume (ESV), and ejection fraction (EF), i.e., the difference between EDV and ESV as the percentage of EDV [3, 4]. As shown in Fig. 1, cardiac sonographers need to segment the LV cavity from the frames at apical 4-chamber (A4C) and apical 2-chamber (A2C) views, determine the long-axis according to the midpoint of the mitral annulus and the apex, then calculate the volume as the sum of a series of parallel slices from apex to base. By strictly following the procedure in the clinical guidelines, clinicians can obtain accurate and reliable quantification results, providing assessments of LV function.

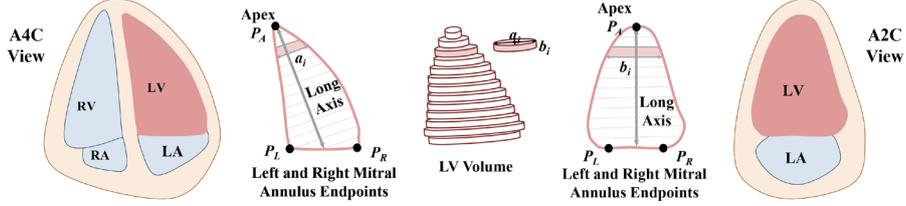

**Fig. 1.** Left ventricular indicators measurements by the biplane Simpson's method.

Many studies have developed automated echocardiography analysis methods to release cardiac sonographers from the laborious work [5, 6], such as numerical regression of LV indices [7, 8], segmentation of LV contours [9-11], and the location of myocardial points [12]. Although deep learning has shown promising results in LV indicator assessments, most existing models are limited by small training data. As a result, it is difficult for algorithms to gain abundant knowledge about general visual representation, hindering further improvement of the LV quantitative performance.

The emerging visual foundation models [13, 14] revolutionized the computer vision field, providing potential solutions to break through the bottleneck. As a representative, the segment anything model (SAM) [15] learns the concept of various objects from numerous images, showing powerful segmentation capacity through simple prompting. Based on SAM, AutoSAMUS [16] designs a parallel CNN branch, a cross-branch attention (CBA) mechanism, and an auto prompt generator (APG), introducing an end-to-end universal model tailored for ultrasound image segmentation. However, AutoSAMUS is mainly oriented to segmentation, which cannot meet the needs of LV indicator measurements in clinical practice. Specifically, after completing LV segmentation, the model has difficulty in directly determining the position of the apex landmark and the mitral annulus from the masks, and then further calculating LV parameters with the biplane Simpson's method recommended by the guidelines.

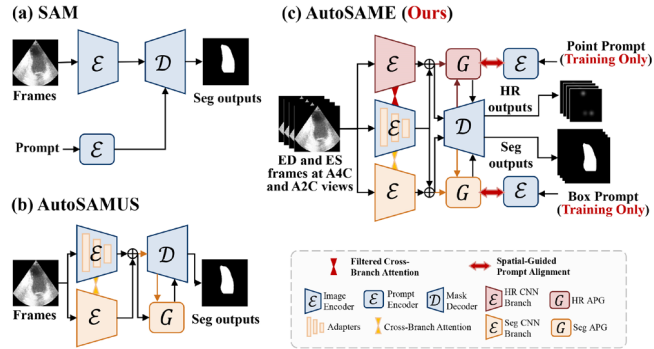

**Fig. 2.** Comparison of (a) SAM, (b) AutoSAMUS, and (c) our AutoSAME. The improvements made by our framework are highlighted in red. By combining SAM's powerful visual understanding with the segmentation and landmark detection task, AutoSAME achieves LV indicators quantification from paired end-diastolic (ED) and end-systolic frames (ES) at A4C and A2C views.

Hence, we propose a novel framework named AutoSAME (where **E** represents **E**chocardiography) to imitate cardiac sonographers to complete LV indicator measurements. **Firstly**, AutoSAME designs two trainable CNN branches with APGs to capture task-specific representations for segmentation versus heatmap regression (HR). Combining the powerful visual understanding of SAM with distinct tasks, AutoSAME delineates the LV contour and locates the key anatomical points simultaneously, quantifying the LV with the biplane Simpson's method, as clinical guidelines recommend. **Secondly**, we advance **filtered cross-branch attention (FCBA)** to decouple segmentation features in the frequency domain and then extend them from the image encoder to the HR CNN branch, providing beneficial information for the HR task. In segmentation, global and local features are often fused to capture the overall scene, e.g., the position relationship between different cardiac chambers, and to perceive edges and textures, such as the boundaries of the myocardium. In contrast, HR tasks typically focus on the approximate region of the key points, which may be beneficial in communication with the segmentation task. Thus, we present FCBA for the adaptive interaction between features from the image encoder with different frequency components and features extracted by the HR CNN branch. **Thirdly**, we introduce the **spatial-guided prompt alignment (SGPA)**, which enhances the reliability and robustness of the generated embeddings by the introduction of prior spatial knowledge. When the coordinates of key anatomical points are projected as embeddings in the prompt encoder, the mask decoder can find shortcuts and output near-perfect landmark localization results. With the prompt encoder as an intermediary, SGPA encourages embeddings from APGs to be consistent with those projected by the prompt encoder based on LV characteristics, improving the quality of APG-generated results.

Our contributions are summarized as follows: 1) For the first time, our novel AutoSAME marries SAM with LV indicators measurements according to clinical guidelines. 2) The advanced FCBA adaptively integrates knowledge from segmentation features to HR features with a frequency perspective, encouraging the latter to receive more comprehensive information. 3) We propose SGPA to generate task-specific prompt embeddings that are more consistent with prior spatial knowledge, improving the performance of LV quantification in an end-to-end manner.

## 2 Method

### 2.1 AutoSAME

**Architecture Overview.** Fig. 2 (c) illustrates our AutoSAME, which aims to marry SAM to left ventricular indicator measurements consistent with clinical guidelines by leveraging its visual comprehension ability in both segmentation and HR tasks. The core components of SAM, i.e., image encoder, prompt encoder, and mask decoder, are preserved to leverage the extensive knowledge accumulated from massive natural images. Referring to the success of SAMUS, we employ position and feature adapters in the image encoder to facilitate the model generalization for echocardiographic analysis and utilize a CNN branch with CBA to supplement local information and an APG to generate prompt embeddings for the segmentation tasks. On this basis, we further introduce the HR CNN branch, HR APG, FCBA and SGPA, which will be detailed below.

**Innovations for Distinct Tasks.** We develop a CNN branch and an APG for landmark localization tasks and present FCBA and SGPA to optimize the visual representation according to the task properties. Since the features extracted by the SAMUS may not be reused for HR tasks, it is necessary to make particular designs to capture features closely related to key point locations in HR, deriving the design of the HR CNN branch with APG. Furthermore, an advanced FCBA transfers multi-level knowledge from the image encoder to the HR CNN branch, while the SGPA imposes constraints on the APG to generate results consistent with the embeddings from the prompt encoder and corresponding to the LV external box or coordinates of key anatomical points, ultimately improving the predictions from the mask decoder.

**Training and Inference.** The inputs of AutoSAME are paired images of A2C and A4C views, including ED and ES frames. Total loss for training is the combination of Dice loss for segmentation, MSE loss for HR, and an alignment loss for SGPA, which will be described in Section 2.3. During inference, the prompt encoder can be completely replaced by SGPA-optimized APG. Moreover, once segmentation masks and coordinates of key anatomical points (the apex point $P_A$, left endpoint $P_L$, and right endpoint $P_R$ of the mitral annulus) are obtained, LV indicator measurements can be performed automatically according to clinical guidelines without manual intervention, as depicted in Fig. 1.

## 2.2   Filtered Cross-Branch Attention

The filtered cross-branch attention (FCBA) realizes the interaction between HR CNN features and image encoder features with different frequency components, dynamically promoting the HR CNN branch to integrate comprehensive knowledge from the image encoder. Since the segmentation naturally drives the model to learn relatively comprehensive features, such as global anatomical structure and local texture details, the expansion of image encoder features can help HR CNN to adaptively select valuable information, improving the learned visual representations from a frequency perspective.

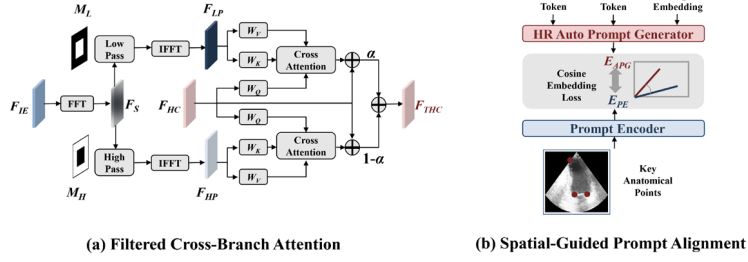

(a) Filtered Cross-Branch Attention    (b) Spatial-Guided Prompt Alignment

**Fig. 3.** The design of (a) FCBA, which promotes the HR CNN branch to dynamically integrate comprehensive knowledge from the image encoder with a frequency domain perspective; and (b) SGPA (take HR branch as the example), which encourages APG to calibrate the generated prompt with spatial prior knowledge of LV with the assistance of prompt encoder outputs.

Fig. 3 (a) details how FCBA optimizes the HR CNN feature with a frequency domain perspective. For a pair of feature maps from the image encoder branch $F_{IE} \in \mathbf{R}^{c \times h \times w}$ and the HR CNN branch $F_{HC} \in \mathbf{R}^{c \times h \times w}$, we first apply a fast Fourier transform (FFT) to $F_{IE}$ and convert it to the spectrum $F_s$. Thereafter, to preserve the specific patterns in $F_{IE}$ corresponding to the grayscale distribution and local texture details, we use low-pass mask $M_L$ and high-pass mask $M_H$ to filter the $F_s$. The masks are defined as follows:

$$M_L \in \mathbf{0}^{h \times w}, M_L\left[\frac{h}{4}:\frac{3h}{4},\frac{w}{4}:\frac{3w}{4}\right] = 1, \quad (1)$$

$$M_H \in \mathbf{1}^{h \times w}, M_H\left[\frac{h}{4}:\frac{3h}{4},\frac{w}{4}:\frac{3w}{4}\right] = 0, \quad (2)$$

after the filtering operation, the feature $F_{LP}$ and $F_{HP}$, according to the low and high frequency components of $F_{IE}$, can be obtained by inverse FFT (IFFT). We employ the learnable matrix $W_Q$ to construct the query by the projective transformation of $F_{HC}$ and utilize $W_K$ and $W_V$ to generate the key-value pairs for both $F_{LP}$ and $F_{HP}$. Then, cross-attentions are adopted for $F_{HC}$ to selectively choose beneficial information from $F_{LP}$ and $F_{HP}$, improving the HR CNN branch's perception ability of the LV shape contours and details around anatomical points in echocardiography. Finally, FCBA executes an adaptive summation through a learnable parameter $\alpha$ to get the tuned HR CNN feature $F_{THC}$, dynamically fusing knowledge of frequency-related patterns for better localizations.

### 2.3  Spatial-Guided Prompt Alignment

The spatial-guided prompt alignment (SGPA) encourages the embeddings generated by the APG to be close to those generated by the prompt encoder based on landmarks and boxes, thereby compensating for the deficiency of APG in exploiting knowledge of LV shape and position, leveraging prior spatial knowledge to improve the reliability and robustness of embeddings.

The way SGPA utilizes spatial prior knowledge to align the prompt embedding is illustrated in Fig. 3(b). Among them, HR APG can automatically generate prompt embedding $E_{APG}$ from the image embedding of the image encoder, the frozen mask token

6     T. Liu et al.

in the mask decoder, and a learnable task token. Meanwhile, the frozen prompt encoder maps the coordinates of key anatomical points to be searched in the LV into embedding $E_{PE}$. Then, we employ the cosine similarity as the alignment loss to push the $E_{APG}$ closer to the $E_{PE}$:

$$L_A = 1 - \cos(E_{APG}, E_{PE}), \qquad (3)$$

where cos represents calculating the cosine of the angle between vectors. SGPA in the segmentation task is identical to that on the HR task, except that the key anatomical points of the input are replaced with the outer bounding box of the LV mask.

## 3  Experiments and Results

### 3.1  Experimental Settings

**Dataset and Evaluation Metrics.** We evaluate our proposed method on the publicly available CAMUS dataset [6], which contains 500 cases with both ED and ES frames at A2C and A4C views. The dataset has provided a manual segmentation mask of the left ventricle, and we further label each image with the location of the apex and two endpoints of the mitral annulus. Model performance is validated with a 10-fold dataset setting (8:1:1 for training, validation and testing), and 5 LV evaluation metrics, i.e. the correlation of EDL, ESL, EDV, ESV, and EF, are included. Besides, we employ the Dice coefficient (DC) and percentage of correct key points (PCK) to quantify the results of segmentation and HR separately.

**Implementation Details.** We run all experiments with Python 3.9 and Pytorch 1.8 environment on an NVIDIA 3090 GPU. Each batch contains 4 images, which are resized to 256 pixels, and then transformed by random rotation, center cropping, and random perspective as augmentation. Adam is employed as the optimization function with a peak learning rate of 0.0002. The epochs are set to 60, where the first 10 epochs undergo a linear warm-up, and the rest is in the cosine decay. The weight ratio of Dice loss to MSE loss is 1:20, and the weight of alignment loss is 1 and is only added in the warm-up phase based on experimental observation. In addition, the standard deviation of the Gaussian heatmap has an initial value of 20 pixels and gradually attenuates to 10 pixels after the warm-up. Additionally, it is noticed that the prompts are no longer needed in testing of AutoSAME.

### 3.2  Ablation study

**Table 1.** The ablation studies validate the effectiveness of the proposed modules in our framework. Compared with the baseline, our AutoSAME improves the prediction of all five LV metrics, such as a 4.3% increase on $EF_{corr}$.

| Methods | FCBA | SGPA | $EDL_{corr}$ | $ESL_{corr}$ | $EDV_{corr}$ | $ESV_{corr}$ | $EF_{corr}$ |
|---|---|---|---|---|---|---|---|
| Baseline(AutoSAMUS) |  |  | 0.898 | 0.877 | 0.949 | 0.951 | 0.784 |
| | √ | | 0.926 | **0.914** | 0.959 | 0.958 | 0.796 |
| | | √ | 0.912 | 0.869 | 0.956 | 0.953 | 0.795 |
| AutoSAME (Ours) | √ | √ | **0.941** | 0.912 | **0.961** | **0.964** | **0.827** |

As tabulated in Table 1, the ablation study results present the effectiveness of each component in AutoSAME for LV indicator measurements compared with the baseline, i.e., AutoSAMUS extended with an HR branch using CBA. FCBA brings a gain of 0.7% to 3.7% to the correlation coefficients of the five LV indicators, which proves the efficacy of information interaction between different branches from the frequency domain perspective. Similarly, the measurement accuracies of the majority of indicators are improved with the introduction of SGPA, implying the positive importance of incorporating prior spatial information into prompt embedding. Moreover, our AutoSAME achieves the best results on most indicators, especially outperforming other settings by at least 3% on the $EF_{corr}$. Besides, the DC of all methods is in the range of 0.927 to 0.929 with subtle differences, while the application of FCBA or SGPA alone improves the PCK (with 1/20 of the input size as the threshold) from 0.928 at the baseline to 0.937 or 0.938, and the combination of the two will bring a gain of 2 percentage points to 0.948.

### 3.3    Comparison with State-of-the-Art Methods

To demonstrate the powerful ability of our framework, in addition to AutoSAMUS, we benchmark the AutoSAME against 7 other state-of-the-art methods across 3 categories, including (1) a multi-task deep learning network for both segmentation and landmark detection tasks named EchoEFNet [17]; (2) four generic vision backbone, as DeepLabV3+ [18], TransUNet [19], SwinUNet [20], and U-Mamba [21]; (3) two pre-trained foundation models tailored for ultrasound image analysis and further fine-tuned on CAMUS, DeblurringMIM [22] and USFM [23].

As shown in Table 2, the proposed AutoSAME gains the best performance compared to all other methods. Moreover, some comparison methods close to AutoSAME on individual metrics are dwarfed by our framework on others. For instance, EchoEFNet, which is closest to AutoSAME on $EDV_{corr}$, misses our framework by 1.9% to 10.5% on other correlation metrics. Due to the careful design of FCBA and SGPA, AutoSAME successfully combines SAM's powerful visual understanding ability with both segmentation and HR tasks, showing comprehensive superiority in LV indicator measurements.

**Table 2.** The quantitative evaluation demonstrates the powerful ability of AutoSAME in segmentation, HR and LV indicator measurements. Our method achieves the best performance on all metrics compared with 7 popular end-to-end methods.

| Methods | DC | PCK | $EDL_{corr}$ | $ESL_{corr}$ | $EDV_{corr}$ | $ESV_{corr}$ | $EF_{corr}$ |
|---|---|---|---|---|---|---|---|
| EchoEFNet[17] | 0.912 | 0.928 | 0.899 | 0.887 | 0.948 | 0.933 | 0.726 |
| DeepLabV3+[18] | 0.904 | 0.917 | 0.866 | 0.830 | 0.912 | 0.881 | 0.619 |
| TransUNet[19] | 0.898 | 0.843 | 0.697 | 0.793 | 0.921 | 0.925 | 0.667 |

| | | | | | | |
|---|---|---|---|---|---|---|
| SwinUNet[20] | 0.917 | 0.935 | 0.929 | 0.888 | 0.959 | 0.942 | 0.810 |
| U-Mamba[21] | 0.911 | 0.902 | 0.864 | 0.835 | 0.942 | 0.939 | 0.814 |
| DeblurringMIM[22] | 0.918 | 0.943 | 0.937 | 0.873 | 0.942 | 0.924 | 0.722 |
| USFM[23] | 0.905 | 0.915 | 0.881 | 0.876 | 0.933 | 0.930 | 0.664 |
| AutoSAME (Ours) | **0.928** | **0.948** | **0.941** | **0.912** | **0.961** | **0.964** | **0.827** |

Visualizations in Fig. 4 further explain that our AutoSAME outperforms all other methods in both segmenting LV contours and locating key anatomical points. On the one hand, benefiting from the strong SAM backbone, our framework suppresses false-positive segmentation mask predictions. On the other hand, the designed FCBA and SGPA extend diverse knowledge to HR tasks to locate the locations of different anatomical points with minimal errors, while most other methods will deviate from the ground truth at the apex. Moreover, LV landmarks and boundaries predicted by AutoSAME can be verifiable mutually. Consequently, our framework is able to quantify the LV indicators accurately from precise segmentation and HR results.

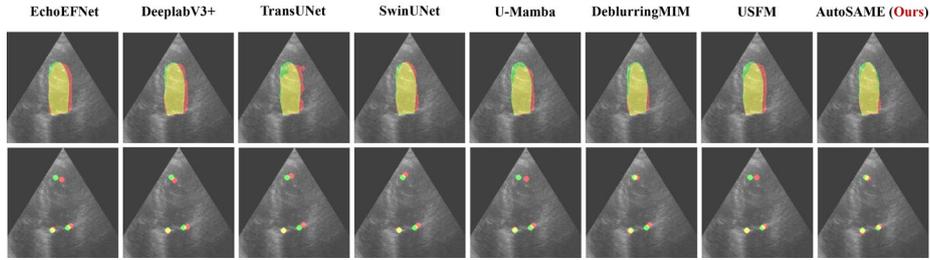

**Fig. 4.** The visualizations explain our AutoSAME's superiority in segmentation and HR compared with 7 methods. Red and green represent the predictions and the ground truth.

## 4    Conclusion

In this paper, we propose AutoSAME, which combines the powerful visual understanding of SAM with both segmentation and landmark localization tasks, achieving the end-to-end LV indicator measurement in a manner consistent with clinical guidelines. Our framework also creatively consists of (1) FCBA, which optimizes the visual representation for landmark localization by integrating comprehensive knowledge from segmentation into HR from a frequency perspective; (2) SGPA, which guides the generation of prompt embeddings by prior spatial knowledge. Extensive experiments with promising results reveal the great clinical potential of our method. In the future, we plan to develop a human-in-the-loop system that applies clinicians' expertise to correct AutoSAME intermediate segmentation and HR results, completing a more efficient and reliable echocardiography analysis process.

## References

1. Hu, H., et al., *A wearable cardiac ultrasound imager.* Nature, 2023. **613**(7945): p. 667-675.

2. Shen, Y., et al., *Development and feasibility of an echocardiography-guided tip location program for central venous catheter implantation.* Nursing in Critical Care, 2025.
3. Mitchell, C., et al., *Guidelines for Performing a Comprehensive Transthoracic Echocardiographic Examination in Adults: Recommendations from the American Society of Echocardiography.* Journal of the American Society of Echocardiography, 2019. **32**(1): p. 1-64.
4. Harkness, A., et al., *Normal reference intervals for cardiac dimensions and function for use in echocardiographic practice: a guideline from the British Society of Echocardiography.* Echo Research and Practice, 2020. **7**(1): p. G1-G18.
5. G, S., U. Gopalakrishnan, R.K. Parthinarupothi, and T. Madathil, *Deep learning supported echocardiogram analysis: A comprehensive review.* Artificial Intelligence in Medicine, 2024. **151**: p. 102866.
6. Leclerc, S., et al., *Deep Learning for Segmentation Using an Open Large-Scale Dataset in 2D Echocardiography.* IEEE Transactions on Medical Imaging, 2019. **38**(9): p. 2198-2210.
7. Kazemi Esfeh, M.M., et al. *A Deep Bayesian Video Analysis Framework: Towards a More Robust Estimation of Ejection Fraction.* 2020. Cham: Springer International Publishing.
8. Muhtaseb, R. and M. Yaqub. *EchoCoTr: Estimation of the Left Ventricular Ejection Fraction from Spatiotemporal Echocardiography.* in *Medical Image Computing and Computer Assisted Intervention – MICCAI 2022.* 2022. Cham: Springer Nature Switzerland.
9. Zhou, G.Q., et al., *DSANet: Dual-Branch Shape-Aware Network for Echocardiography Segmentation in Apical Views.* IEEE Journal of Biomedical and Health Informatics, 2023. **27**(10): p. 4804-4815.
10. Ding, W., et al., *Multiple token rearrangement Transformer network with explicit superpixel constraint for segmentation of echocardiography.* Medical Image Analysis, 2025. **101**: p. 103470.
11. Lin, J., W. Xie, L. Kang, and H. Wu, *Dynamic-Guided Spatiotemporal Attention for Echocardiography Video Segmentation.* IEEE Transactions on Medical Imaging, 2024. **43**(11): p. 3843-3855.
12. Azad, M.A., et al. *EchoTracker: Advancing Myocardial Point Tracking in Echocardiography.* 2024. Cham: Springer Nature Switzerland.
13. Awais, M., et al., *Foundation Models Defining a New Era in Vision: a Survey and Outlook.* IEEE Transactions on Pattern Analysis and Machine Intelligence, 2025: p. 1-20.
14. He, Y., et al., *Foundation Model for Advancing Healthcare: Challenges, Opportunities and Future Directions.* Ieee Reviews in Biomedical Engineering, 2025. **18**: p. 172-191.
15. Kirillov, A., et al. *Segment Anything.* in *2023 IEEE/CVF International Conference on Computer Vision (ICCV).* 2023.
16. Lin, X., Y. Xiang, L. Yu, and Z. Yan. *Beyond Adapting SAM: Towards End-to-End Ultrasound Image Segmentation via Auto Prompting.* 2024. Cham: Springer Nature Switzerland.
17. Li, H., et al., *EchoEFNet: Multi-task deep learning network for automatic calculation of left ventricular ejection fraction in 2D echocardiography.* Computers in Biology and Medicine, 2023. **156**: p. 106705.
18. Chen, L.-C., et al. *Encoder-Decoder with Atrous Separable Convolution for Semantic Image Segmentation.* 2018. Cham: Springer International Publishing.
19. Chen, J., et al., *TransUNet: Transformers Make Strong Encoders for Medical Image Segmentation.* arXiv preprint arXiv:2102.04306.
20. Cao, H., et al. *Swin-Unet: Unet-Like Pure Transformer for Medical Image Segmentation.* 2023. Cham: Springer Nature Switzerland.

21. Ma, J., F. Li, and B. Wang, *U-Mamba: Enhancing Long-range Dependency for Biomedical Image Segmentation.* arXiv preprint arXiv:2401.04722.
22. Kang, Q., et al., *Deblurring masked image modeling for ultrasound image analysis.* Medical Image Analysis, 2024. **97**: p. 103256.
23. Jiao, J., et al., *USFM: A universal ultrasound foundation model generalized to tasks and organs towards label efficient image analysis.* Medical Image Analysis, 2024. **96**: p. 103202.